\documentclass[lettersize,journal]{IEEEtran}
\usepackage{amsmath,amsfonts}
\usepackage{algorithmic}
\usepackage{algorithm}
\usepackage{array}
\usepackage[caption=false,font=footnotesize,labelfont=sf,textfont=sf]{subfig}
\usepackage{textcomp}
\usepackage{stfloats}
\usepackage{url}
\usepackage{verbatim}
\usepackage{graphicx}
\usepackage{cite}

\usepackage{booktabs} 
\usepackage{graphicx} 
\usepackage{amssymb}
\usepackage{multirow}

\begin{document}

\title{AgentGate: A Lightweight Structured Routing Engine for the Internet of Agents}

\author{
  Yujun Cheng$^{1}$,~\IEEEmembership{Member,~IEEE,} Enfang Cui$^{2}$, Hao Qin$^{1}$, Zhiyuan Liang$^{2}$, Qi Xu$^{3}$ \\
  $^{1}$School of Artificial Intelligence, University of Science and Technology Beijing, Beijing, China \\
  $^{2}$China Telecom Research Institute, China \\
  $^{3}$Hangzhou Institute for Advanced Study, University of Chinese Academy of Sciences, Hangzhou, China \\
  Emails: yjcheng@ustb.edu.cn, \{cuief, liangzy17\}@chinatelecom.cn, m202520907@xs.ustb.edu.cn, xuqi@bjtu.edu.cn

  \thanks{This paper was produced by the IEEE Publication Technology Group. They are in Piscataway, NJ.}
  \thanks{Manuscript received April 19, 2021; revised August 16, 2021.}
}

\markboth{Journal of \LaTeX\ Class Files,~Vol.~14, No.~8, August~2021}%
{Cheng \MakeLowercase{\textit{et al.}}: AgentGate: A Lightweight Structured Routing Engine for the Internet of Agents}


\maketitle

\begin{abstract}
  The rapid development of AI agent systems is leading to an emerging Internet of Agents, where specialized agents operate across local devices, edge nodes, private services, and cloud platforms. Although recent efforts have improved agent naming, discovery, and interaction, efficient request dispatch remains an open systems problem under latency, privacy, and cost constraints. In this paper, we present AgentGate, a lightweight structured routing engine for candidate-aware agent dispatch. Instead of treating routing as unrestricted text generation, AgentGate formulates it as a constrained decision problem and decomposes it into two stages: action decision and structural grounding. The first stage determines whether a query should trigger single-agent invocation, multi-agent planning, direct response, or safe escalation, while the second stage instantiates the selected action into executable outputs such as target agents, structured arguments, or multi-step plans. To adapt compact models to this setting, we further develop a routing-oriented fine-tuning scheme with candidate-aware supervision and hard negative examples. Experiments on a curated routing benchmark with several 3B--7B open-weight models show that compact models can provide competitive routing performance in constrained settings, and that model differences are mainly reflected in action prediction, candidate selection, and structured grounding quality. These results indicate that structured routing is a feasible design point for efficient and privacy-aware agent systems, especially when routing decisions must be made under resource-constrained deployment conditions.
\end{abstract}

\begin{IEEEkeywords}
  AI Agents, Edge Intelligence, Multi-Agent Systems, Structured Routing, AgentDNS
\end{IEEEkeywords}

\section{Introduction}
With the rapid rise of AI agent systems, we are beginning to witness the emergence of an Internet of Agents \cite{chen2024internet}, in which large numbers of specialized agents interact, collaborate, and provide services across local devices, edge nodes, private clusters, and cloud platforms \cite{cheng2024toward,liu2026joint,zhang2025resource}. In such an ecosystem, foundational infrastructure is gradually taking shape. Agent naming and discovery mechanisms, exemplified by systems such as AgentDNS \cite{cui2025agentdns}, address how agents can be registered and located, while protocols such as A2H \cite{liang2025a2h} define how humans and agents exchange tasks and responses. However, once an agent request is issued, the system must still decide whether the request should be handled locally or remotely, whether it should be delegated to a single agent or decomposed into multiple steps, and whether execution should proceed at all under latency, privacy, and safety constraints.

This agent routing problem is particularly important in edge intelligence settings. In many practical deployments, the agent routing engine itself is most naturally placed on the user side or the edge side, where it can observe requests early, reduce unnecessary cloud exposure, and make low-latency dispatch decisions. Yet the candidate agents being routed to are not necessarily edge-resident, some may be lightweight local tools or private services, while others may be remote APIs or cloud-based expert agents. This creates a fundamental systems tension. Relying on a powerful cloud LLM as the default router can improve generality, but it also places every request on the critical path of a remote model, introducing extra latency, repeated API cost, and broader data exposure \cite{zhang2024llm}. Conversely, relying only on a compact local model improves privacy and efficiency, but may struggle on ambiguous or long-tail routing cases. What is therefore missing is a structured routing engine that can operate locally for routine requests while supporting principled escalation when stronger external reasoning is needed.

Existing work on multi-agent systems \cite{wu2024autogen,qian2024chatdev,hong2023metagpt} has primarily focused on agent collaboration, planning, and tool use, often assuming that a strong centralized model orchestrates the entire process. While effective in open-ended settings, this paradigm is suboptimal for frequent dispatch decisions in edge environments. In practice, many requests do not require full-scale reasoning by a frontier cloud model. Instead, they require a constrained decision over a candidate agent set: whether to invoke one agent, generate a multi-agent plan, return a direct response, or stop execution due to safety or authorization concerns. From this perspective, routing should not be treated as unrestricted text generation. Rather, it should be formulated as a structured decision problem with explicit action boundaries and executable outputs.

Motivated by this observation, we propose AgentGate, a structured routing engine for AI agents. AgentGate is designed to run at the edge and to make routing decisions under practical deployment constraints, including latency, privacy, and heterogeneous backend availability. Instead of directly generating a complete routing result in one pass, AgentGate decomposes routing into two stages. The first stage performs action decision, selecting among single-agent invocation, multi-agent planning, direct response, and safe escalation. The second stage performs structural grounding, instantiating the selected action into executable outputs such as a target agent, schema-conformant arguments, or an ordered multi-agent plan. This design makes the routing process easier to constrain, interpret, and validate, especially when the router is implemented using compact 3B–7B class models \cite{deng2025crosslm}. To adapt compact models to this structured routing setting, we further employ a candidate-aware fine-tuning strategy tailored to agent registries and constrained outputs. The training data cover single-agent invocation, multi-agent planning, direct fallback, and safe escalation, together with hard negatives to improve routing boundary discrimination. In addition, although AgentGate is designed for edge-side deployment, its confidence-aware design also supports selective fallback to a stronger remote model for difficult routing cases. Besides, we evaluate AgentGate on a curated routing benchmark using several 3B–7B open-weight backbones. Results show that compact models can achieve strong routing performance under structured supervision, while different model families exhibit complementary strengths across action prediction, candidate selection, argument grounding, and plan generation. These findings indicate that structured routing is a practical and promising component for emerging AI agent ecosystems.

The main contributions of this paper are summarized as follows:

\begin{itemize}
  \item We formulate structured routing for AI agents as a candidate-aware decision problem, in which the router must choose among single-agent invocation, multi-agent planning, direct response, and safe escalation under explicit output constraints.

  \item We propose AgentGate, a lightweight two-stage routing engine that decomposes routing into action decision and structural grounding, enabling executable and interpretable routing with compact edge-side models.

  \item We develop a candidate-aware fine-tuning pipeline with routing-oriented supervision and hard negative design, and show through experiments on 3B–7B open-weight models that small models can serve as effective structured routers for AI agent ecosystems.
\end{itemize}

The rest of this paper is organized as follows. Section~II reviews related work on multi-agent systems, tool use, and edge model adaptation. Section~III formulates the routing problem. Section~IV presents the proposed AgentGate architecture and training methodology. Section~V reports experimental results and analysis. Section~VI concludes the paper and discusses future directions.

\begin{figure}[t]
  \centering
  \includegraphics[width=3.4in]{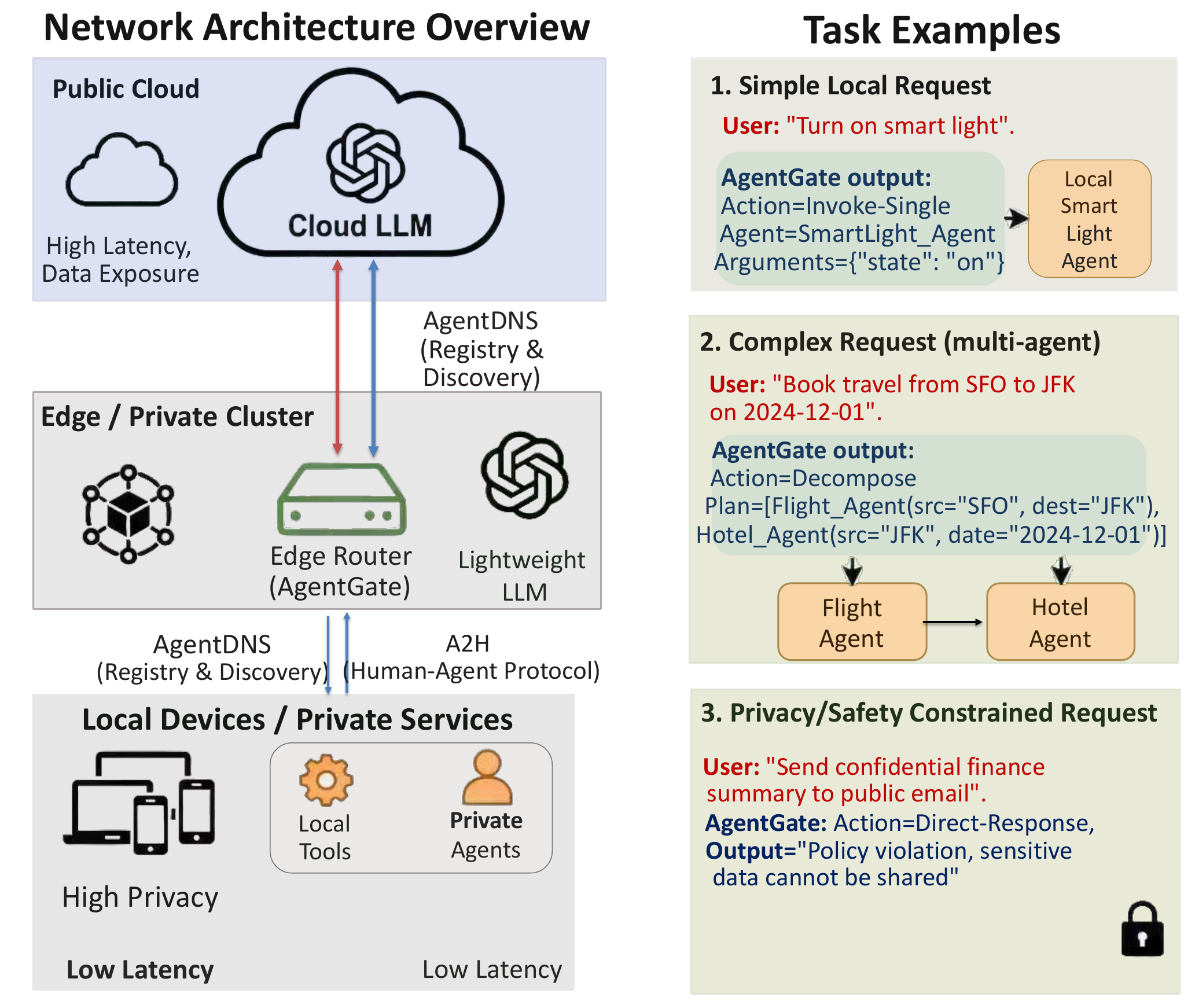}
  \caption{Overview of the proposed agent network architecture and representative routing task examples handled by AgentGate.}
  \label{fig:main}
\end{figure}

\section{Related Work}

\subsection{Multi-Agent Systems and Agent Infrastructure}
Recent advances in large language models (LLMs) have catalyzed the development of multi-agent systems (MAS), enabling complex task resolution through agent collaboration \cite{wu2024autogen,hong2023metagpt}. As the Internet of Agents paradigm emerges, foundational infrastructures are actively being proposed. For instance, AgentDNS mechanisms address agent registration and discovery \cite{cui2025agentdns}, while interaction protocols like A2H standardize user-agent exchanges \cite{liang2025a2h}. However, the critical intermediate layer of agent routing remains largely underexplored. Existing MAS frameworks generally assume a static collaboration topology or rely on centralized, unrestricted communication \cite{qian2024chatdev,tran2025multi}. In contrast, AgentGate explicitly addresses the dynamic routing problem, namely determining whether a request should be processed locally, delegated to a specific agent, decomposed, or rejected, which is essential for practical, resource-constrained deployments.

\subsection{LLM-Based Orchestration and Tool Use}
Extensive research has explored LLMs as central orchestrators capable of tool invocation, function calling, and structured generation \cite{schick2023toolformer,patil2024gorilla,shen2024hugginggpt}. While constrained decoding improves execution reliability \cite{li2025adaptrack}, conventional tool-use paradigms treat the LLM as the primary problem solver that utilizes tools merely as auxiliary aids \cite{xu2025llm}. Furthermore, these systems typically assume a relevant tool must be invoked whenever possible. AgentGate fundamentally differs by formulating routing as a structured decision problem. Rather than acting as a general solver, our router serves as a specialized dispatcher. It explicitly supports critical non-invocation actions, such as direct fallback, safe escalation, and execution abstention, making it more suitable for real-world scenarios bounded by safety and authorization constraints.

\subsection{Edge Intelligence and Small Language Models}
The demand for low-latency, privacy-preserving AI has accelerated the deployment of edge intelligence \cite{cheng2024rcif, cheng2025snapcfl, cheng2025seful}. Because frontier cloud models are often too resource-intensive for edge devices, recent efforts have focused on adapting Small Language Models (SLMs) via model compression and parameter-efficient fine-tuning (e.g., LoRA) \cite{hu2022lora,dettmers2023qlora}. While SLMs demonstrate strong performance on narrowly defined tasks with aligned supervision \cite{touvron2023llama,jiang2023mistral7b}, their application as edge-side routing engines for MAS has not been thoroughly investigated. AgentGate bridges this gap. By employing candidate-aware fine-tuning and structured action spaces, we demonstrate that compact 3B to 7B class models can effectively execute candidate-aware, safe, and executable routing decisions without over-relying on massive cloud-based controllers.

\section{Problem Formulation}

In this section, we formalize agent routing in an edge multi-agent environment. Our goal is to map a natural-language query and a candidate agent set to a structured routing decision that can be executed reliably by downstream components.

\subsection{System Setting}

Consider an edge agent network with a local registry of available agents, denoted by $\mathcal{C} = \{c_1, c_2, \dots, c_N\}$. Each agent $c_i \in \mathcal{C}$ is associated with a metadata tuple $c_i = \langle n_i, d_i, \mathcal{A}_i \rangle$, where $n_i$ denotes the agent name, $d_i$ is a textual description of its capability, and $\mathcal{A}_i$ specifies the argument schema required for invocation.

Given a user query $q$, a lightweight retrieval module first selects a candidate subset $\mathcal{C}_{\mathrm{sub}} \subseteq \mathcal{C}$ that contains agents potentially relevant to the query. The router then receives an input tuple $x = \langle q, \mathcal{C}_{\mathrm{sub}}, s \rangle$, where $s$ denotes optional contextual information such as conversation history or system state.

In practice, $\mathcal{C}$ is not assumed to be a globally complete agent set. Instead, it denotes the registry accessible to the edge router at inference time. This accessible registry may be formed from locally cached agent metadata and, when necessary, augmented by external AgentDNS-based discovery. Accordingly, the candidate subset $\mathcal{C}_{\mathrm{sub}}$ is defined over the accessible registry rather than over the full ecosystem-wide agent space. Under this setting, the routing task is defined as a local decision problem conditioned on the currently accessible candidate set. The router is not intended to solve the user query itself in the general case; rather, it predicts an appropriate routing action and, when needed, produces a structured execution output consumable by downstream agents or controllers.

\subsection{Constrained Action Space}

To support practical deployment in edge environments, we define routing as a decision process over a constrained action space $\mathcal{Y} = \{y_{\mathrm{call}}, y_{\mathrm{plan}}, y_{\mathrm{direct}}, y_{\mathrm{escalate}}\}$, where each action has the following semantics:

\begin{itemize}
  \item \textbf{Single-agent invocation} ($y_{\mathrm{call}}$): the query can be handled by invoking one target agent in $\mathcal{C}_{\mathrm{sub}}$.
  \item \textbf{Multi-agent planning} ($y_{\mathrm{plan}}$): the query requires multiple agents to be executed in sequence or coordination.
  \item \textbf{Direct response} ($y_{\mathrm{direct}}$): the router returns a direct natural-language response instead of invoking any registered agent. This includes conversational fallback or cases where no suitable candidate exists.
  \item \textbf{Safe escalation} ($y_{\mathrm{escalate}}$): the query is rejected or deferred due to safety, privacy, authorization, or policy-related concerns.
\end{itemize}

This formulation differs from conventional open-ended tool use in that routing explicitly includes \emph{non-invocation decisions}. In practical edge settings, forcing every query into an agent call may lead to invalid execution, privacy leakage, or unsafe behavior. The inclusion of $y_{\mathrm{direct}}$ and $y_{\mathrm{escalate}}$ therefore provides an explicit mechanism for boundary-aware dispatch.

\subsection{Structured Routing Output}

Given an input $x$, the router produces a structured output through a routing model $\mathcal{F}_{\theta}$ parameterized by $\theta$:
\begin{equation}
  o = \mathcal{F}_{\theta}(x).
\end{equation}

We represent the routing output as
\begin{equation}
  o = \langle y, \hat{c}, \Omega, \Pi, r, \gamma \rangle,
  \label{eq:routing_output}
\end{equation}
where $y \in \mathcal{Y}$ is the predicted routing action, $\hat{c} \in \mathcal{C}_{\mathrm{sub}} \cup \{\varnothing\}$ is the selected target agent, $\Omega$ denotes the structured argument set for invoking $\hat{c}$, $\Pi$ is an ordered multi-agent execution plan, $r$ is an optional textual rationale or response content, and $\gamma \in [0,1]$ is a confidence score associated with the routing decision.

The internal structure of the plan can be written as
\begin{equation}
  \Pi = \big[(\hat{c}^{(1)}, \Omega^{(1)}), (\hat{c}^{(2)}, \Omega^{(2)}), \dots, (\hat{c}^{(T)}, \Omega^{(T)})\big],
  \label{eq:plan}
\end{equation}
where each pair $(\hat{c}^{(t)}, \Omega^{(t)})$ denotes the selected agent and argument set at step $t$.

Not all fields in Eq.~(\ref{eq:routing_output}) are active for every action. When $y = y_{\mathrm{call}}$, the output mainly consists of $\hat{c}$ and $\Omega$. When $y = y_{\mathrm{plan}}$, the plan $\Pi$ becomes the primary executable object. When $y \in \{y_{\mathrm{direct}}, y_{\mathrm{escalate}}\}$, no agent invocation is produced, and the output primarily contains the action label together with the textual response or rationale.

To make this dependency explicit, the executable portion of the output can be viewed as action-dependent:
\begin{equation}
  \mathrm{Exec}(o)=
  \begin{cases}
    (\hat{c}, \Omega), & y = y_{\mathrm{call}},                                \\[4pt]
    \Pi,               & y = y_{\mathrm{plan}},                                \\[4pt]
    \varnothing,       & y \in \{y_{\mathrm{direct}}, y_{\mathrm{escalate}}\}.
  \end{cases}
  \label{eq:exec}
\end{equation}

This formulation highlights that routing is not merely a classification problem. Instead, it requires the model to jointly determine \emph{what action should be taken} and \emph{what executable structure should be produced} under that action.

\subsection{Routing Objective}

Given a training set $\mathcal{D} = \{(x^{(k)}, o^{(k)})\}_{k=1}^{K}$, the goal is to learn a routing model $\mathcal{F}_{\theta}$ that maximizes the conditional likelihood of the target structured outputs:
\begin{equation}
  \max_{\theta} \sum_{k=1}^{K} \log p_{\theta}\!\left(o^{(k)} \mid x^{(k)}\right).
  \label{eq:overall_objective}
\end{equation}

However, the routing task is inherently compositional. The router must first determine the coarse action type, then identify the appropriate candidate agent or execution plan, and finally generate arguments that conform to the expected schema. This motivates the following factorization:
\begin{equation}
  p_{\theta}(o \mid x)
  =
  p_{\theta}(y \mid x)\;
  p_{\theta}(\hat{c}, \Omega, \Pi, r, \gamma \mid x, y).
  \label{eq:factorization}
\end{equation}

Eq.~(\ref{eq:factorization}) reflects a natural separation between \emph{action prediction} and \emph{structural grounding}. The first term determines the routing intent under the constrained action space, while the second term instantiates the decision into executable outputs conditioned on the predicted action.

From this perspective, agent routing is more appropriately viewed as a structured decision problem with explicit output constraints, rather than a purely free-form generation task. This formulation motivates the two-stage routing architecture introduced in the next section.

\begin{figure*}[htbp]
  \centering
  \includegraphics[width=6.0in]{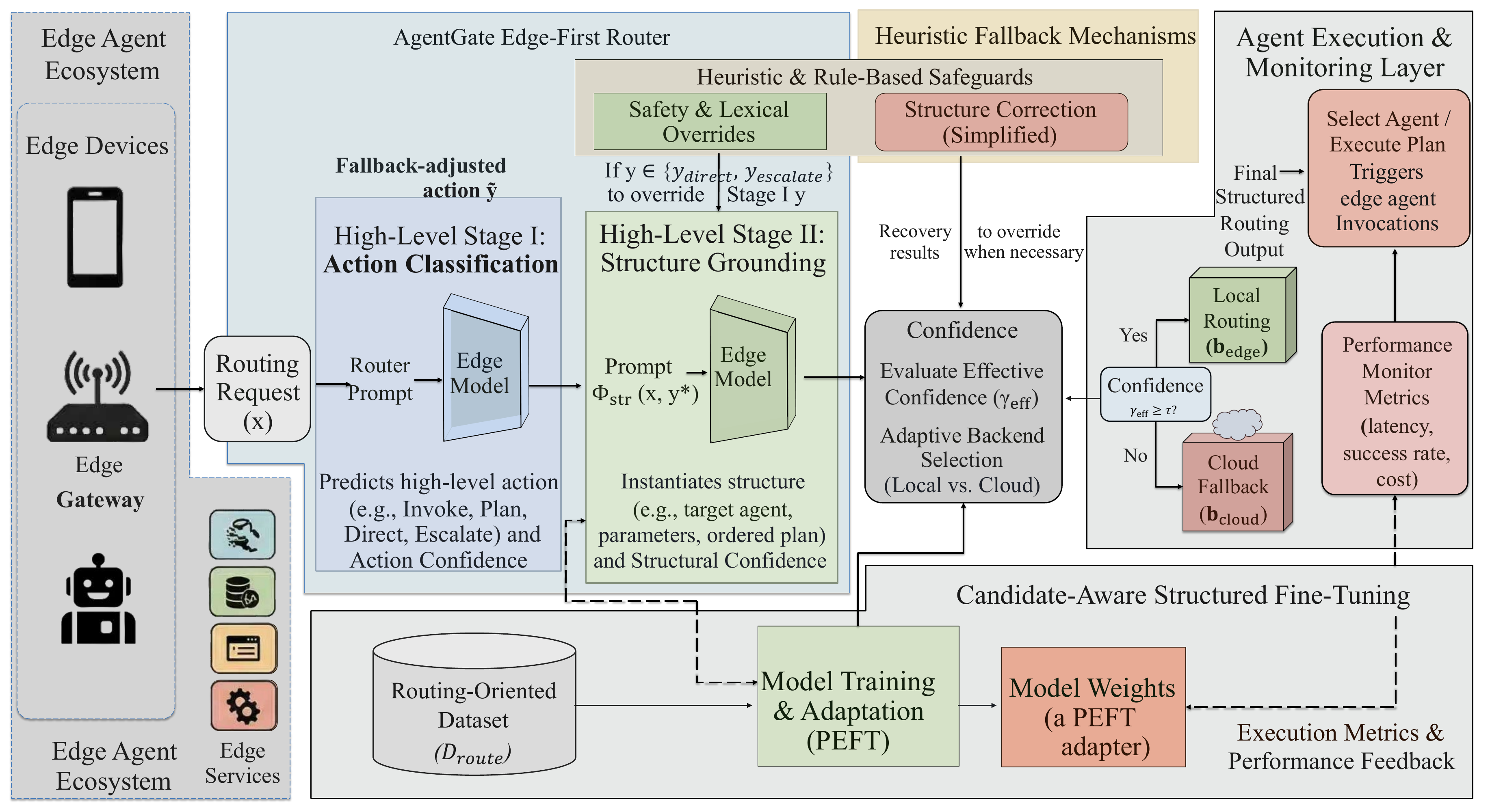}
  \caption{Detailed architecture of the AgentGate framework, illustrating the two-stage structured routing pipeline, heuristic fallback mechanisms, and confidence-aware backend selection.}
  \label{fig:cross_column}
\end{figure*}

\section{Methodology}

To address the structured routing problem defined in Section III, we propose AgentGate, an edge-first and confidence-aware routing framework for edge agent networks. Rather than treating routing as unrestricted one-shot generation, AgentGate first attempts routing locally using a lightweight two-stage router, and selectively defers low-confidence cases to a stronger remote model. Within each backend, routing is decomposed into action decision and structural grounding, which together produce an executable structured routing output.

\subsection{Overview of the Confidence-Aware Two-Stage Router}

To address the structured routing problem defined in Section III, we propose AgentGate, an edge-first and confidence-aware routing framework for edge agent networks. The core idea is to perform routing locally by default, while allowing low-confidence cases to be selectively deferred to a stronger remote model when necessary. Within each routing backend, AgentGate decomposes routing into two sequential subproblems: a coarse action decision stage and a conditional structural grounding stage. This design reduces the difficulty of generating a complete routing output in a single pass and makes the routing process easier to constrain, interpret, and stabilize under limited model capacity.

Let $x = \langle q, \mathcal{C}_{\mathrm{sub}}, s \rangle$ denote the routing input. Instead of directly generating the full structured output $o = \langle y, \hat{c}, \Omega, \Pi, r, \gamma \rangle$ in one step, AgentGate factorizes the routing process into two stages:

\begin{equation}
  p_{\theta}(o \mid x) = p_{\theta}(y, r, \gamma_{\mathrm{act}} \mid x)\; p_{\theta}(\hat{c}, \Omega, \Pi, \gamma_{\mathrm{str}} \mid x, y).
  \label{eq:two_stage_factorization}
\end{equation}

The first term models \emph{action decision}, namely whether the query should trigger single-agent invocation, multi-agent planning, direct response, or safe escalation. The second term models \emph{structural grounding}, where the selected action is instantiated into executable outputs such as a target agent, a plan, or schema-conformant arguments.

This decomposition is motivated by the observation that action prediction and structure generation require different capabilities. The former mainly concerns coarse intent discrimination and boundary awareness, whereas the latter requires fine-grained candidate selection and parameter grounding. By separating them, the router can focus on a smaller decision space at each stage.

AgentGate adopts an edge-first routing workflow. Concretely, the local edge model first attempts to solve the routing problem through the two-stage process above. Let $\gamma_{\mathrm{act}}$ and $\gamma_{\mathrm{str}}$ denote the confidence scores produced by the action decision and structural grounding stages, respectively. We define the effective routing confidence as
\begin{equation}
  \gamma_{\mathrm{eff}} = \min(\gamma_{\mathrm{act}}, \gamma_{\mathrm{str}}).
\end{equation}

For cases that terminate after the first stage, such as $y_{\mathrm{direct}}$ or $y_{\mathrm{escalate}}$, we take $\gamma_{\mathrm{eff}} = \gamma_{\mathrm{act}}$.

Based on the effective confidence, the routing backend is selected by
\begin{equation}
  b^{*} =
  \begin{cases}
    b_{\mathrm{edge}},  & \gamma_{\mathrm{eff}} \ge \tau, \\
    b_{\mathrm{cloud}}, & \gamma_{\mathrm{eff}} < \tau,
  \end{cases}
\end{equation}

\noindent where $b_{\mathrm{edge}}$ denotes local routing with the edge-side small model, $b_{\mathrm{cloud}}$ denotes fallback to a stronger remote model, and $\tau$ is a predefined confidence threshold.

This backend decision should be distinguished from the routing action $y_{\mathrm{escalate}}$. The action $y_{\mathrm{escalate}}$ is a task-level decision indicating that the query should not proceed through normal execution due to safety, privacy, authorization, or policy constraints. By contrast, the backend choice in Eq.~(9) is an inference-level decision that determines which model computes the routing decision itself. In other words, safe escalation controls whether a request should proceed, whereas backend fallback controls how the routing decision is computed.

Overall, this design yields a unified framework with three desirable properties. First, the two-stage decomposition improves routing controllability by separating action prediction from executable grounding. Second, the edge-first workflow preserves the latency, privacy, and cost advantages of local routing for routine cases. Third, confidence-triggered cloud fallback allows difficult routing instances to benefit from stronger remote reasoning without forcing all requests through the cloud.

\subsection{Stage I: Action Decision}

In the first stage, the router predicts an action $y \in \mathcal{Y}$ together with an optional rationale $r$ and an action confidence score $\gamma_{\mathrm{act}}$. Let $\Phi_{\mathrm{act}}(\cdot)$ denote the stage-specific prompt template for action prediction. The first-stage output is defined as

\begin{equation}
  (y, r, \gamma_{\mathrm{act}})
  =
  \mathcal{F}_{\theta}^{\mathrm{act}}\!\big(\Phi_{\mathrm{act}}(x)\big).
  \label{eq:stage1}
\end{equation}

The action space is constrained to $\mathcal{Y} = \{y_{\mathrm{call}}, y_{\mathrm{plan}}, y_{\mathrm{direct}}, y_{\mathrm{escalate}}\}$, which encourages the model to make an explicit routing decision before attempting detailed grounding. This stage acts as an early filter: if the predicted action is $y_{\mathrm{direct}}$ or $y_{\mathrm{escalate}}$, the pipeline can terminate immediately without invoking the second stage.

To make this decision process explicit, the first-stage policy can be written as
\begin{equation}
  y^{*}
  =
  \arg\max_{y \in \mathcal{Y}}
  p_{\theta}(y \mid x).
  \label{eq:action_decision}
\end{equation}

This formulation allows the model to distinguish between executable requests and boundary cases before generating agent-level outputs. In practice, this is important for reducing unnecessary downstream computation and for improving routing reliability in the presence of unsupported, ambiguous, or sensitive queries.

\subsection{Stage II: Structural Grounding}

If the predicted action is executable, i.e., $y^{*} \in \{y_{\mathrm{call}}, y_{\mathrm{plan}}\}$, the router proceeds to the second stage. Let $\Phi_{\mathrm{str}}(\cdot)$ denote the structural grounding template conditioned on both the input and the stage-one action. The second-stage generation is defined as
\begin{equation}
  (\hat{c}, \Omega, \Pi, \gamma_{\mathrm{str}})
  =
  \mathcal{F}_{\theta}^{\mathrm{str}}\!\big(\Phi_{\mathrm{str}}(x, y^{*})\big).
  \label{eq:stage2}
\end{equation}

When $y^{*} = y_{\mathrm{call}}$, the model selects one target agent $\hat{c} \in \mathcal{C}_{\mathrm{sub}}$ and generates its corresponding argument set $\Omega$. When $y^{*} = y_{\mathrm{plan}}$, the model outputs an ordered execution plan
\begin{equation}
  \Pi = \big[(\hat{c}^{(1)}, \Omega^{(1)}), (\hat{c}^{(2)}, \Omega^{(2)}), \dots, (\hat{c}^{(T)}, \Omega^{(T)})\big],
  \label{eq:plan_generation}
\end{equation}
where each pair $(\hat{c}^{(t)}, \Omega^{(t)})$ specifies the selected agent and argument instantiation at step $t$.

The stage-two decision can be interpreted as a constrained grounding problem:
\begin{equation}
  (\hat{c}, \Omega, \Pi)^{*}
  =
  \arg\max_{\hat{c}, \Omega, \Pi}
  p_{\theta}(\hat{c}, \Omega, \Pi \mid x, y^{*}),
  \label{eq:grounding_decision}
\end{equation}
subject to candidate consistency and schema validity. In other words, the generated structure must remain compatible with the predicted action, the available candidate agents, and their invocation schemas.

Compared with one-shot routing, this conditional design reduces the search space of the second stage. Once the action has been fixed, the model no longer needs to jointly consider incompatible alternatives such as direct response versus agent invocation, which helps stabilize the generation of executable outputs.

\subsection{Heuristic Fallback Mechanisms}

Although the two-stage router is learned end-to-end, structured routing in practice still benefits from lightweight deterministic safeguards. We therefore incorporate fallback mechanisms to handle stage failures, invalid outputs, and incomplete grounding results.

\subsubsection{Action-Side Safeguards}

Let $\mathcal{W}_{\mathrm{sens}}$ denote a predefined set of sensitivity-related lexical or semantic cues, and let $\mathcal{W}_{\mathrm{seq}}$ denote a set of sequential markers such as ``first,'' ``then,'' or ``after that.'' Based on these cues, we define a fallback-adjusted action $\tilde{y}$ as
\begin{equation}
  \tilde{y} =
  \begin{cases}
    y_{\mathrm{escalate}}, & \text{if } q \text{ contains sensitive or unauthorized content},                     \\[4pt]
    y_{\mathrm{plan}},     & \text{if } q \text{ sequential composition and } |\mathcal{C}_{\mathrm{sub}}| \ge 2, \\[4pt]
    y_{\mathrm{call}},     & \text{if stage I fails and } \mathcal{C}_{\mathrm{sub}} \neq \varnothing,            \\[4pt]
    y_{\mathrm{direct}},   & \text{if stage I fails and } \mathcal{C}_{\mathrm{sub}} = \varnothing,               \\[4pt]
    y^{*},                 & \text{otherwise.}
  \end{cases}
  \label{eq:action_fallback}
\end{equation}

This safeguard is not intended to replace the learned model. Instead, it provides a conservative control layer that improves robustness when the first stage produces malformed outputs or encounters obvious boundary cases.

\subsubsection{Candidate Recovery and Slot Completion}

If stage II fails to produce a valid target agent or an executable plan, we apply a lightweight candidate recovery rule based on metadata matching. For each candidate $c_i \in \mathcal{C}_{\mathrm{sub}}$, we define a heuristic relevance score
\begin{equation}
  S(c_i, q)
  =
  \sum_{w \in q} \mathbf{1}\!\big(w \in T(c_i)\big)
  +
  \lambda \sum_{h \in \mathcal{H}}
  \mathbf{1}\!\big(h \in T(c_i)\big)\mathbf{1}(h \in q),
  \label{eq:heuristic_score}
\end{equation}
where $T(c_i)$ denotes the tokenized metadata of candidate $c_i$, $\mathcal{H}$ is a set of manually defined semantic hints, and $\lambda$ is a weighting factor. The candidate with the highest score is selected as a fallback target:
\begin{equation}
  \hat{c}_{\mathrm{fb}} = \arg\max_{c_i \in \mathcal{C}_{\mathrm{sub}}} S(c_i, q).
  \label{eq:fallback_candidate}
\end{equation}

If the generated argument set is incomplete, we further apply a lightweight slot-completion module to extract values directly from the query. Let $\mathcal{E}_{\mathrm{slot}}(q, \mathcal{A}_{\hat{c}})$ denote this extractor conditioned on the target schema. The completed argument set is written as
\begin{equation}
  \Omega_{\mathrm{fb}} = \mathcal{E}_{\mathrm{slot}}(q, \mathcal{A}_{\hat{c}}).
  \label{eq:slot_completion}
\end{equation}

In our implementation, this module can be instantiated with regular expressions or other lightweight extraction rules for common slot types such as time, location, entity name, or numeric values. The purpose is not to replace learned grounding, but to improve executability when the model output is partially correct but structurally incomplete.

\begin{algorithm}[t]
  \caption{AgentGate Confidence-Aware Structured Routing}
  \label{alg:agentgate_routing}
  \begin{algorithmic}[1]
    \REQUIRE User query $q$, candidate registry subset $\mathcal{C}_{sub}$, optional context $s$, edge routing model $\mathcal{F}_{edge}$, cloud fallback model $\mathcal{F}_{cloud}$, and confidence threshold $\tau$.
    \ENSURE Executable structured routing output $o = \langle \bar{y}, \hat{c}, \Omega, \Pi, r \rangle$.

    \STATE Initialize input tuple $x = \langle q, \mathcal{C}_{sub}, s \rangle$.
    \STATE Set active routing model $\mathcal{F} \leftarrow \mathcal{F}_{edge}$.
    \WHILE{True}
    \STATE \textbf{/* Stage I: Action Decision */}
    \STATE Predict initial action, rationale, and confidence: $(y, r, \gamma_{act}) \leftarrow \mathcal{F}^{act}(\Phi_{act}(x))$
    \STATE Apply safety \& lexical heuristics to obtain fallback-adjusted action $\bar{y}$ (Eq. 15)

    \IF{$\bar{y} \in \{y_{direct}, y_{escalate}\}$}
    \STATE Set effective confidence $\gamma_{eff} \leftarrow \gamma_{act}$
    \STATE Set structural execution outputs $(\hat{c}, \Omega, \Pi) \leftarrow (\emptyset, \emptyset, \emptyset)$
    \ELSE
    \STATE \textbf{/* Stage II: Structural Grounding */}
    \STATE Generate structural instantiation: $(\hat{c}, \Omega, \Pi, \gamma_{str}) \leftarrow \mathcal{F}^{str}(\Phi_{str}(x, \bar{y}))$

    \IF{target agent $\hat{c}$ is invalid or missing}
    \STATE Recover candidate via metadata matching: $\hat{c} \leftarrow \arg\max_{c_i \in \mathcal{C}_{sub}} S(c_i, q)$ (Eq. 17)
    \ENDIF

    \IF{argument set $\Omega$ is incomplete}
    \STATE Apply slot completion: $\Omega \leftarrow \Omega \cup \mathcal{E}_{slot}(q, \mathcal{A}_{\hat{c}})$ (Eq. 18)
    \ENDIF

    \STATE Calculate effective confidence: $\gamma_{eff} \leftarrow \min(\gamma_{act}, \gamma_{str})$
    \ENDIF

    \STATE \textbf{/* Adaptive Backend Selection */}
    \IF{$\gamma_{eff} < \tau$ \AND $\mathcal{F} == \mathcal{F}_{edge}$}
    \STATE Trigger cloud fallback: set $\mathcal{F} \leftarrow \mathcal{F}_{cloud}$
    \STATE \textbf{continue}
    \ELSE
    \STATE \textbf{break}
    \ENDIF
    \ENDWHILE

    \RETURN Final structured output $o = \langle \bar{y}, \hat{c}, \Omega, \Pi, r \rangle$.
  \end{algorithmic}
\end{algorithm}

\subsection{Candidate-Aware Structured Fine-Tuning}

General instruction-tuning data are not naturally aligned with routing tasks, where outputs must satisfy action consistency, candidate awareness, and schema constraints. To adapt small language models to this setting, we construct a routing-oriented supervised fine-tuning dataset
\begin{equation}
  \mathcal{D}_{\mathrm{route}}
  =
  \big\{(x^{(k)}, o^{(k)})\big\}_{k=1}^{K},
  \label{eq:dataset}
\end{equation}
where each example contains a query, a candidate agent subset, optional context, and a structured target output.

The training data are organized into four categories:
\begin{itemize}
  \item \textbf{Single-agent invocation}: queries that can be resolved by exactly one candidate agent;
  \item \textbf{Multi-agent planning}: compositional queries requiring multi-step execution across agents;
  \item \textbf{Direct fallback}: conversational or unsupported queries for which no invocation should be made;
  \item \textbf{Safe escalation}: sensitive, unauthorized, or policy-violating requests that should not be executed locally.
\end{itemize}

To sharpen decision boundaries, we further introduce hard negatives during data construction. These include cases with misleading sequential wording, semantically overlapping candidates, and ambiguous requests that resemble executable queries but should instead trigger direct response or escalation. The goal is to prevent the model from overfitting to superficial lexical cues and to improve discrimination among closely related routing decisions.

Let $\Phi_{\mathrm{train}}(x)$ denote the serialized prompt containing the query, candidate metadata, and optional context. The model is trained to generate the target structured output autoregressively:
\begin{equation}
  \mathcal{L}_{\mathrm{SFT}}
  =
  -
  \sum_{t=1}^{T}
  \log p_{\theta}\!\big(o_t \mid o_{<t}, \Phi_{\mathrm{train}}(x)\big),
  \label{eq:sft_loss}
\end{equation}
where $o_t$ denotes the $t$-th token in the target routing output.

To encourage structural specialization, the loss is computed only over target output tokens rather than over the full prompt. This design focuses learning on action labels, candidate references, plans, and argument fields, thereby aligning the model more directly with the structured routing objective.

In practice, we implement this adaptation using parameter-efficient fine-tuning so that the backbone model remains lightweight enough for edge-oriented deployment. The resulting router combines the flexibility of pretrained language models with task-specific structural supervision tailored to candidate-aware agent routing.

\subsection{Confidence-Aware Hybrid Edge-Cloud Routing}

Although AgentGate is primarily designed as a lightweight edge-side router, the proposed architecture also supports a confidence-aware hybrid deployment mode. The motivation is practical: many routing queries can be handled efficiently by a local small language model, but a small fraction of ambiguous or long-tail cases may still benefit from stronger remote reasoning capability. Instead of routing all queries through a cloud model, AgentGate allows the backend to be selected adaptively according to routing confidence.

Let $\gamma_{\mathrm{act}}$ and $\gamma_{\mathrm{str}}$ denote the confidence scores produced by Stage~I and Stage~II, respectively. We define the effective routing confidence as
\begin{equation}
  \gamma_{\mathrm{eff}} = \min(\gamma_{\mathrm{act}}, \gamma_{\mathrm{str}}).
  \label{eq:effective_confidence}
\end{equation}
For cases that terminate after Stage~I, such as $y_{\mathrm{direct}}$ or $y_{\mathrm{escalate}}$, $\gamma_{\mathrm{eff}}$ is taken as $\gamma_{\mathrm{act}}$.

Based on this score, the routing backend can be selected by a simple thresholding rule:
\begin{equation}
  b^{*} =
  \begin{cases}
    b_{\mathrm{edge}},  & \gamma_{\mathrm{eff}} \ge \tau, \\[4pt]
    b_{\mathrm{cloud}}, & \gamma_{\mathrm{eff}} < \tau,
  \end{cases}
  \label{eq:backend_selection}
\end{equation}
where $b_{\mathrm{edge}}$ denotes local routing with the edge-side small model, $b_{\mathrm{cloud}}$ denotes fallback to a stronger remote API model, and $\tau$ is a predefined confidence threshold.

This mechanism should be distinguished from the routing action $y_{\mathrm{escalate}}$. The action $y_{\mathrm{escalate}}$ is a task-level decision indicating that a query should not be executed locally due to safety, authorization, or policy considerations. By contrast, the backend choice in Eq.~(\ref{eq:backend_selection}) is an inference-level decision that determines which model performs the routing itself. In other words, safe escalation controls \emph{whether} a request should proceed, whereas backend escalation controls \emph{how} the routing decision is computed.

In practice, this design provides a flexible trade-off between efficiency and robustness. Routine and well-formed routing requests can be resolved locally with low latency and limited data exposure, while difficult cases can be selectively delegated to a stronger remote model only when necessary. Although the present work focuses on the edge-side routing setting, this hybrid mode makes the overall framework easier to extend to more heterogeneous deployment environments.

\section{Experiments}

This section evaluates AgentGate in terms of routing accuracy,
diagnostic behavior, and deployment characteristics. We begin with the
benchmark and implementation setup, then report the main comparisons
with alternative routing paradigms and backbone families. The final
parts examine the effect of task-specific adaptation and the runtime
properties of the system, including hybrid edge--cloud execution.

\subsection{Experimental Setup}

This subsection describes the benchmark construction, task definition,
evaluation metrics, and implementation settings used throughout the
experiments.

\subsubsection{Benchmark and Task Definition}

We construct a curated runtime benchmark containing 3,200 AgentDNS
\cite{cui2025agentdns} routing instances, split into 2,400 training, 400 validation, and 400
test samples. To prevent the task from collapsing into superficial
lexical matching, the dataset includes hard negative cases such as
semantically overlapping candidate tools, deceptive sequential queries,
and sensitive escalation triggers.

The benchmark targets candidate-aware multi-agent routing at the
application layer. Each instance contains a user query, a candidate
set of available agents, and optional contextual metadata. The model
must generate a structured routing decision from a constrained action
space consisting of \texttt{CALL\_AGENT},
\texttt{MULTI\_AGENT\_PLAN}, \texttt{DIRECT\_ANSWER}, and
\texttt{ESCALATE}. For executable cases, the output must also be
grounded into agent-specific arguments or a multi-step plan over the
candidate set.

The benchmark spans a range of service domains, including food
delivery, ride hailing, lodging, weather, restaurant booking, grocery
ordering, movie tickets, courier services, flight booking, and
calendar management. This setting is intended to approximate runtime
conditions in which multiple partially overlapping agents may be
available at the same time.

\subsubsection{Evaluation Metrics}

We report action accuracy, agent accuracy, argument exact match, plan
exact match, JSON validity, and escalation precision and recall. These
metrics cover three aspects of the problem: coarse routing accuracy,
structural grounding quality, and safety-related escalation behavior.

\subsubsection{Backbones and Training Details}

The evaluated backbones are representative open-weight models in the
3B--7B range, including Qwen2.5-3B, Qwen2.5-7B, Mistral-7B,
Llama2-7B, and Phi-3.5-mini. To match edge-oriented deployment
constraints, all fine-tuned models are adapted with LoRA-based
parameter-efficient fine-tuning on a single NVIDIA GeForce RTX 5090
GPU.

\subsection{Main Results}

This subsection reports the main empirical comparisons: first against
alternative routing paradigms, and then across backbone families under
the AgentGate framework.

\subsubsection{Comparison with Alternative Routing Paradigms}

To assess whether the AgentGate formulation is necessary, we compare it
with three alternative paradigms: rule-based heuristics,
retrieve-rank pipelines, and generic LLM tool calling. These baselines
test whether the AgentDNS routing task can be handled by simpler
lexical rules, retrieval-oriented matching, or off-the-shelf tool
invocation, without the candidate-aware two-stage decomposition used
in AgentGate.

The \textbf{Rule-based} baseline uses lexical triggers, metadata
matching, and handcrafted safeguards to determine routing actions and
fill arguments. This baseline serves as a lower bound, especially
because our system itself includes heuristic fallback components. The
\textbf{Retrieve-rank} baseline first performs lightweight action
classification, then ranks candidate agents using semantic and lexical
query--agent matching, followed by simple slot completion. The
\textbf{Tool-Calling} baseline treats candidate agents as ordinary
callable tools and prompts the LLM to emit a tool invocation directly,
without explicit action decomposition or structural grounding.

\begin{figure*}[t]
  \centering
  \includegraphics[width=4.5in]{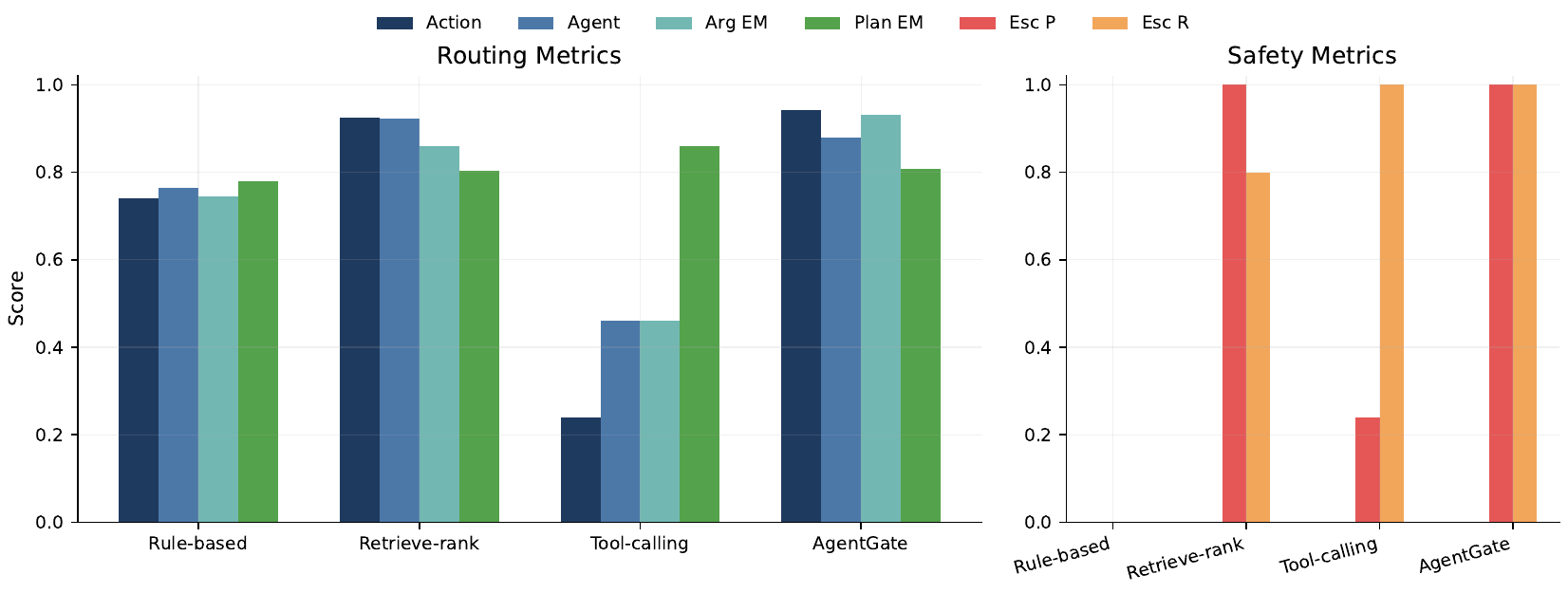}
  \caption{Baseline comparison on the AgentDNS runtime benchmark. The
    left panel reports routing metrics, and the right panel reports
    escalation precision and recall.}
  \label{fig:baseline-overview}
\end{figure*}

Table~\ref{tab:baseline-comparison} and
Figure~\ref{fig:baseline-overview} summarize the results. The
rule-based baseline remains syntactically valid, but its performance
on the main routing metrics is clearly lower, and escalation handling
is unreliable. Lexical rules alone are therefore insufficient for
reliable AgentDNS routing.

The retrieve-rank pipeline is a meaningful baseline. It reaches
competitive results on action prediction, agent selection, and
argument grounding, which indicates that a substantial portion of the
benchmark can be handled through strong candidate matching. Even so,
AgentGate remains better on overall structured routing quality and on
safety-sensitive behavior.

Generic tool calling performs substantially worse than AgentGate,
despite perfect JSON validity. This gap indicates that AgentDNS
routing is not equivalent to standard function invocation. The
comparison makes two points clear: retrieval-style matching is
helpful, but heuristic rules are too brittle, and generic tool calling
does not replace explicit structured routing. The two-stage design of
AgentGate is therefore retained in the remainder of the study.

\begin{table*}[t]
  \centering
  \small
  \setlength{\tabcolsep}{5pt}
  \begin{tabular}{lcccccc}
    \toprule
    Method                    & Action          & Agent  & Arg EM          & Plan EM         & Esc P           & Esc R           \\
    \midrule
    Rule-based                & 0.7400          & 0.7650 & 0.7450          & 0.7800          & 0.0000          & 0.0000          \\
    Retrieve-rank             & 0.9250          & 0.9225 & 0.8600          & 0.8025          & 1.0000          & 0.8000          \\
    Tool-Calling (Qwen2.5-7B) & 0.2400          & 0.4600 & 0.4600          & 0.8600          & 0.2400          & 1.0000          \\
    AgentGate (Qwen2.5-7B)    & \textbf{0.9425} & 0.8800 & \textbf{0.9325} & \textbf{0.8075} & \textbf{1.0000} & \textbf{1.0000} \\
    \bottomrule
  \end{tabular}
  \caption{Comparison with alternative routing paradigms on the
    \texttt{agentdns\_runtime} benchmark. Retrieve-rank is a competitive
    baseline, but AgentGate achieves better overall routing quality.}
  \label{tab:baseline-comparison}
\end{table*}

\subsubsection{Backbone Comparison under the AgentGate Framework}

Having established the value of the routing formulation, we next
compare different backbone families under the same AgentGate setting.
The task definition, training data, and routing pipeline are kept
fixed, so that the effect of the underlying language model can be
examined directly.

Table~\ref{tab:backbone-comparison} reports the results. Among the
evaluated backbones, \textbf{Qwen2.5-7B} gives the most balanced
result set across routing quality and safety behavior. It reaches
high action accuracy and argument exact match, while also achieving
perfect escalation precision and recall on this benchmark.

\textbf{Qwen2.5-3B} is noticeably weaker, especially in agent
selection and escalation recall. This leaves room for lightweight
deployment, but with a visible reduction in routing quality.
\textbf{Phi-3.5-mini} is also highly competitive and attains the best
action and agent accuracy among the evaluated models. Its advantage is
less pronounced from a deployment perspective, however, because its
runtime behavior is slower and less stable in our system
measurements.

By contrast, \textbf{Mistral-7B} and \textbf{Llama2-7B} are less
reliable in the runtime setting, particularly on structured grounding
and escalation-related metrics. The shared routing formulation does
not eliminate backbone effects. Performance still depends on how well
the underlying model handles constrained decisions, candidate-aware
grounding, and structured JSON generation. Among the evaluated
backbones, Qwen2.5-7B offers the most favorable trade-off between
routing quality, safety behavior, and deployment cost.

\begin{table*}[t]
  \centering
  \small
  \setlength{\tabcolsep}{5pt}
  \begin{tabular}{lcccccc}
    \toprule
    Backbone     & Action          & Agent           & Arg EM          & Plan EM         & Esc P           & Esc R           \\
    \midrule
    Qwen2.5-3B   & 0.8975          & 0.7025          & 0.6925          & 0.8075          & 0.9231          & 0.4800          \\
    Qwen2.5-7B   & 0.9425          & 0.8800          & \textbf{0.9325} & \textbf{0.8075} & \textbf{1.0000} & \textbf{1.0000} \\
    Phi-3.5-mini & \textbf{0.9625} & \textbf{0.9300} & 0.8600          & 0.8025          & \textbf{1.0000} & 0.8000          \\
    Mistral-7B   & 0.6500          & 0.2425          & 0.3500          & 0.8000          & \textbf{1.0000} & 0.8000          \\
    Llama2-7B    & 0.4500          & 0.5075          & 0.5675          & 0.8025          & 0.2475          & \textbf{1.0000} \\
    \bottomrule
  \end{tabular}
  \caption{Backbone comparison under the AgentGate routing framework on
    the \texttt{agentdns\_runtime} benchmark. Qwen2.5-7B gives the most
    balanced result set across routing and escalation metrics.}
  \label{tab:backbone-comparison}
\end{table*}

\subsection{Ablation and Diagnostic Analysis}

This subsection examines which factors contribute to the behavior of
AgentGate, with an emphasis on task-specific adaptation.

\subsubsection{Effect of Task-Specific Fine-Tuning}

To clarify the role of task adaptation, we compare fine-tuned and
non-fine-tuned models under the same two-stage routing framework. The
routing pipeline, candidate set, and benchmark subset are fixed, and
the only varying factor is whether the backbone is adapted with
task-specific supervision. This comparison is intended as a diagnostic
analysis rather than a primary leaderboard result.

Table~\ref{tab:fine-tuned-vs-base} reports the results on a two-stage
probe. Fine-tuning does not consistently improve top-level action
prediction. For several strong backbones, the base model already
achieves high action and agent accuracy under the two-stage prompting
setup. This means that modern instruction-tuned LLMs already carry
substantial coarse routing priors when the candidate set is made
explicit.

The effect of fine-tuning appears more clearly in
\textbf{argument exact match}. In several cases, the fine-tuned models
produce outputs that are more schema-aligned and more regular in
format, whereas the base models are more likely to produce values that
are semantically reasonable but structurally mismatched. The gain from
task-specific supervision is therefore concentrated more on structured
grounding fidelity than on coarse intent recognition.

That gain is not uniform across backbones. For some models,
fine-tuning improves argument exactness or output regularity without
improving action or agent prediction; for others, the effect is more
mixed. Task-specific supervision mainly affects schema alignment and
output regularity, rather than uniformly improving all routing
metrics.

\begin{table}[t]
  \centering
  \small
  \setlength{\tabcolsep}{5pt}
  \begin{tabular}{llccc}
    \toprule
    Backbone   & Setting    & Action & Agent  & Arg EM \\
    \midrule
    Qwen2.5-3B & Base       & 0.9531 & 0.9688 & 0.8750 \\
    Qwen2.5-3B & Fine-tuned & 0.8906 & 0.9219 & 0.8906 \\
    \midrule
    Qwen2.5-7B & Base       & 0.9844 & 0.9844 & 0.9219 \\
    Qwen2.5-7B & Fine-tuned & 0.9844 & 0.9375 & 0.8906 \\
    \midrule
    Mistral-7B & Base       & 0.9844 & 0.9844 & 0.8438 \\
    Mistral-7B & Fine-tuned & 0.9844 & 0.9375 & 0.8906 \\
    \midrule
    Llama2-7B  & Base       & 0.9844 & 0.9375 & 0.8906 \\
    Llama2-7B  & Fine-tuned & 0.9844 & 0.9688 & 0.9063 \\
    \bottomrule
  \end{tabular}
  \caption{Fine-tuned versus base models on two-stage
    routing probe. The main differences appear in argument exact match
    rather than in coarse routing accuracy.}
  \label{tab:fine-tuned-vs-base}
\end{table}

\subsection{System Study}

This subsection focuses on deployment-related behavior, including
local runtime efficiency and hybrid edge--cloud execution.

\subsubsection{System Performance and Latency}

Beyond routing accuracy, AgentGate is intended to operate as a runtime
decision system. We therefore measure end-to-end route latency,
action-stage latency, selection-stage latency, throughput,
time-to-first-token (TTFT), and GPU memory usage. All measurements are
collected under offline non-streaming local inference on a single GPU
with the same routing pipeline and benchmark subset. Table~\ref{tab:system-performance} reports the main system metrics. One pattern is consistent across the main
backbones: the selection stage accounts for most of the
runtime cost. Under the current implementation, the second-stage
structural grounding module is the main source of latency in the
two-stage design.

Efficiency also varies substantially across backbones. Although
Qwen2.5-7B is larger than Qwen2.5-3B, it is slightly faster in average
route latency under our implementation, while also delivering higher
routing accuracy. Llama2-7B, by comparison, incurs much higher
latency, which makes it considerably less attractive for deployment.

TTFT and decoding throughput offer a more fine-grained view of system
behavior. For the Qwen family, TTFT remains modest relative to total
route latency, and output throughput is broadly stable. The dominant
overhead therefore comes less from token generation itself and more
from repeated structured reasoning and grounding steps inside the
routing pipeline.

\begin{figure*}[t]
  \centering
  \subfloat[Throughput--latency trade-off across representative backbones.
    Qwen2.5-7B combines lower route latency with higher throughput under
    the same measurement setup.\label{fig:qps-vs-latency}]{
    \includegraphics[width=2in]{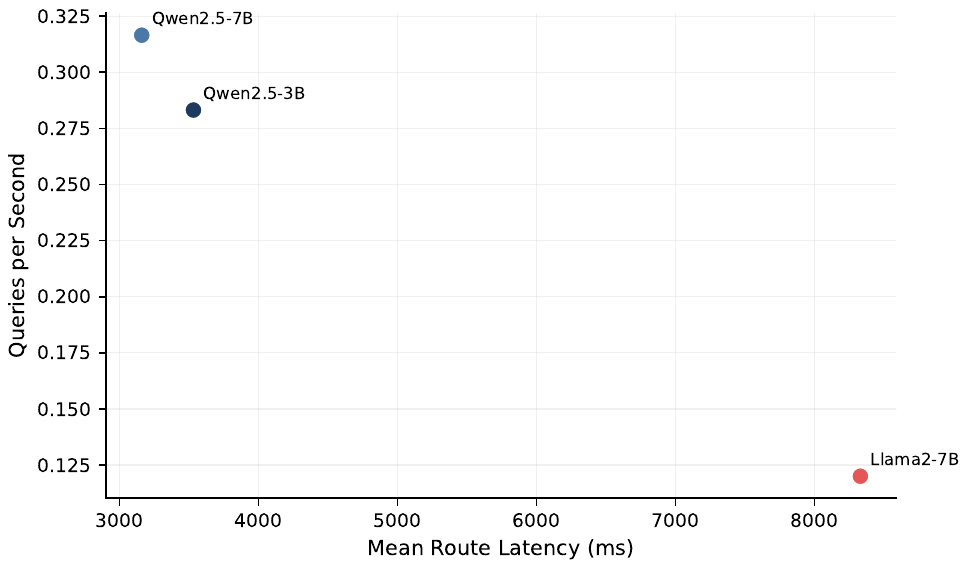}
  }
  \quad \quad
  \subfloat[Memory--latency trade-off across representative backbones.
    Qwen2.5-3B uses the least memory, whereas Qwen2.5-7B offers a better
    latency--quality trade-off.\label{fig:memory-vs-latency}]{
    \includegraphics[width=2in]{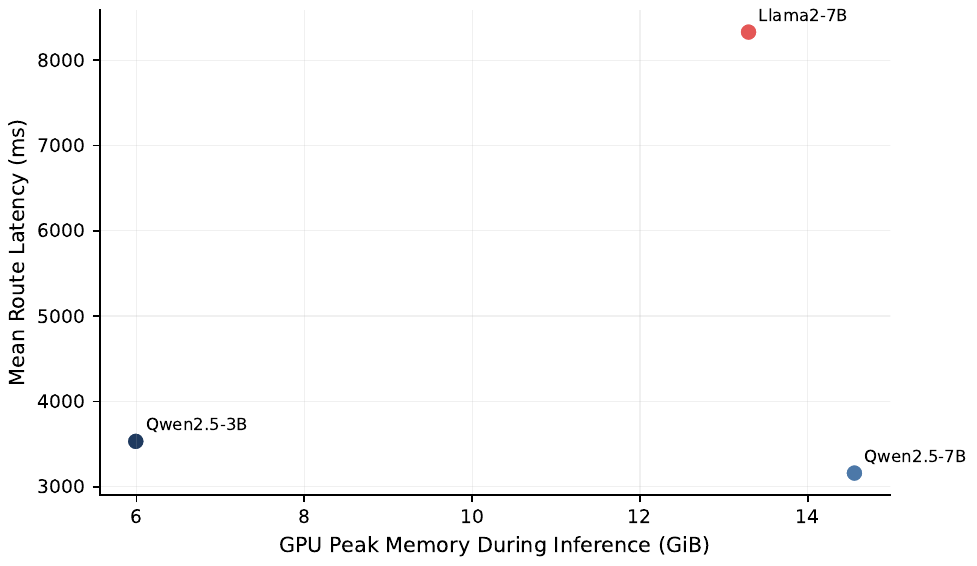}
  }
  \caption{Performance trade-offs across representative backbones.}
  \label{fig:combined-tradeoffs}
\end{figure*}

Figure~\ref{fig:qps-vs-latency} plots the throughput--latency trade-off
across representative backbones. Qwen2.5-7B occupies the most
favorable region, combining lower route latency with higher
throughput than Qwen2.5-3B and Llama2-7B.
Figure~\ref{fig:memory-vs-latency} shows the memory--latency trade-off.
Qwen2.5-3B has the smallest memory footprint, whereas Qwen2.5-7B
offers a better balance between latency and routing quality.

These measurements place AgentGate on the systems side of the problem,
rather than treating it only as an accuracy benchmark. In the current
implementation, the second-stage grounding module is the main runtime
cost, and Qwen2.5-7B gives the most favorable balance between routing
quality and deployment efficiency among the measured backbones.

\begin{table*}[t]
  \centering
  \small
  \setlength{\tabcolsep}{4pt}
  \begin{tabular}{lccccc}
    \toprule
    Model      & QPS   & Route Lat. (ms) & Action Lat. (ms) & Selection Lat. (ms) & GPU Mem. (GiB) \\
    \midrule
    Qwen2.5-3B & 0.283 & 3531.5          & 709.2            & 3762.8              & 5.99           \\
    Qwen2.5-7B & 0.317 & 3159.5          & 594.0            & 3420.4              & 14.56          \\
    Llama2-7B  & 0.120 & 8329.8          & 4745.5           & 4778.9              & 13.29          \\
    \bottomrule
  \end{tabular}
  \caption{System performance on a 20-sample runtime probe for the
    backbones with directly comparable two-stage measurements.}
  \label{tab:system-performance}
\end{table*}

\subsubsection{Hybrid Edge--Cloud Routing}

A design goal of AgentGate is to support confidence-aware hybrid
routing, in which low-confidence edge decisions can be deferred to a
stronger cloud model. This setting is motivated by the need to balance
routing quality, latency, and deployment cost. In our framework, the
edge model first produces a structured routing decision together with
confidence signals, after which a fallback policy may invoke a
stronger cloud backend when the local decision appears unreliable.

Table~\ref{tab:hybrid-routing} reports the real hybrid results on a
50-sample probe. Under the current forced-fallback configuration, the
hybrid setting largely collapses to the cloud-only line for several
backbones. Qwen2.5-7B, Qwen2.5-3B, and Mistral-7B, for example, yield
nearly identical hybrid scores. Once fallback is triggered
aggressively, final performance is dominated by the cloud backend
rather than by the edge model.

This experiment should be read mainly as a feasibility check rather
than as evidence that the current hybrid policy is already
Pareto-optimal. The routing chain itself works end to end, but the
main remaining bottleneck lies in fallback triggering and confidence
calibration. The issue is no longer whether cloud fallback can be
integrated, but whether the system can decide more effectively when
fallback should be invoked.

\begin{table*}[t]
  \centering
  \small
  \setlength{\tabcolsep}{4pt}
  \begin{tabular}{lcccccc}
    \toprule
    Setting              & Action          & Agent           & Arg EM          & Plan EM & Esc P           & Esc R           \\
    \midrule
    Qwen2.5-7B edge-only & \textbf{1.0000} & \textbf{0.9400} & \textbf{0.9400} & 0.8600  & \textbf{1.0000} & \textbf{1.0000} \\
    Qwen2.5-7B hybrid    & 0.9800          & \textbf{0.9400} & 0.9000          & 0.8600  & \textbf{1.0000} & 0.9167          \\
    Qwen2.5-3B hybrid    & 0.9800          & \textbf{0.9400} & 0.9000          & 0.8600  & \textbf{1.0000} & 0.9167          \\
    Mistral-7B hybrid    & 0.9800          & \textbf{0.9400} & 0.9000          & 0.8600  & \textbf{1.0000} & 0.9167          \\
    Llama2-7B hybrid     & 0.9000          & 0.8600          & 0.8200          & 0.8600  & 0.7692          & 0.8333          \\
    \bottomrule
  \end{tabular}
  \caption{Real edge--cloud routing results. The
    cloud backend uses MiniMax M2.5.
    Under forced fallback, most hybrid results approach the cloud-only
    line.}
  \label{tab:hybrid-routing}
\end{table*}

\section{Conclusion}

In this paper, we proposed AgentGate, a lightweight structured agent routing engine for Agent network. By formulating agent routing as a structured decision problem, AgentGate decomposes the routing process into two stages: action decision and structural grounding. This design enables the router to predict whether a query should trigger single-agent invocation, multi-agent planning, direct fallback, or safe escalation, and then generate the corresponding executable structure.

Experimental results on a curated AgentDNS benchmark show that small language models can achieve effective routing performance under candidate-aware supervision and explicit structural constraints. The results also suggest that strict output constraints and hard negative design are important for improving routing reliability. Future work will extend this framework to broader and more dynamic agent environments, and will further evaluate end-to-end execution performance and system-level efficiency under realistic edge deployment settings.

\bibliographystyle{IEEEtran}
\bibliography{aaai2026}

\vfill

\end{document}